\documentclass[letterpaper]{article} 
\usepackage{aaai2026}  
\usepackage{times}  
\usepackage{helvet}  
\usepackage{courier}  
\usepackage[hyphens]{url}  
\usepackage{graphicx} 
\usepackage{booktabs}
\usepackage{natbib}  
\usepackage{caption} 
\usepackage{amsmath}
\usepackage{amsfonts}
\usepackage{multirow}
\usepackage{algorithm}
\usepackage{algorithmic}
\usepackage{amssymb}   
\usepackage{booktabs}  
\usepackage{makecell}  
\usepackage{newfloat}
\usepackage{listings}
\DeclareCaptionStyle{ruled}{labelfont=normalfont,labelsep=colon,strut=off} 
\floatstyle{ruled}
\newfloat{listing}{tb}{lst}{}
\floatname{listing}{Listing}
\usepackage{bm}

\title{SkeNa: Learning to Navigate Unseen Environments\\Based on Abstract Hand-Drawn Maps}
\author {
    Haojun Xu\textsuperscript{\rm 1},
    Jiaqi Xiang\textsuperscript{\rm 1},
    Wei Wu\textsuperscript{\rm 1},
    Linqing Zhong\textsuperscript{\rm 1},
    Jinyu Chen\textsuperscript{\rm 1},
    Linjiang Huang\textsuperscript{\rm 1}\thanks{Corresponding author.},
    Hongyu Yang\textsuperscript{\rm 1}\thanks{Corresponding author.},
    Si Liu\textsuperscript{\rm 1}\thanks{Corresponding author.}
}
\affiliations {
    \textsuperscript{\rm 1}Beihang University
}

\usepackage{bibentry}

\begin{document}

\maketitle
\begin{figure*}
    \centering
    \includegraphics[width=1\linewidth]{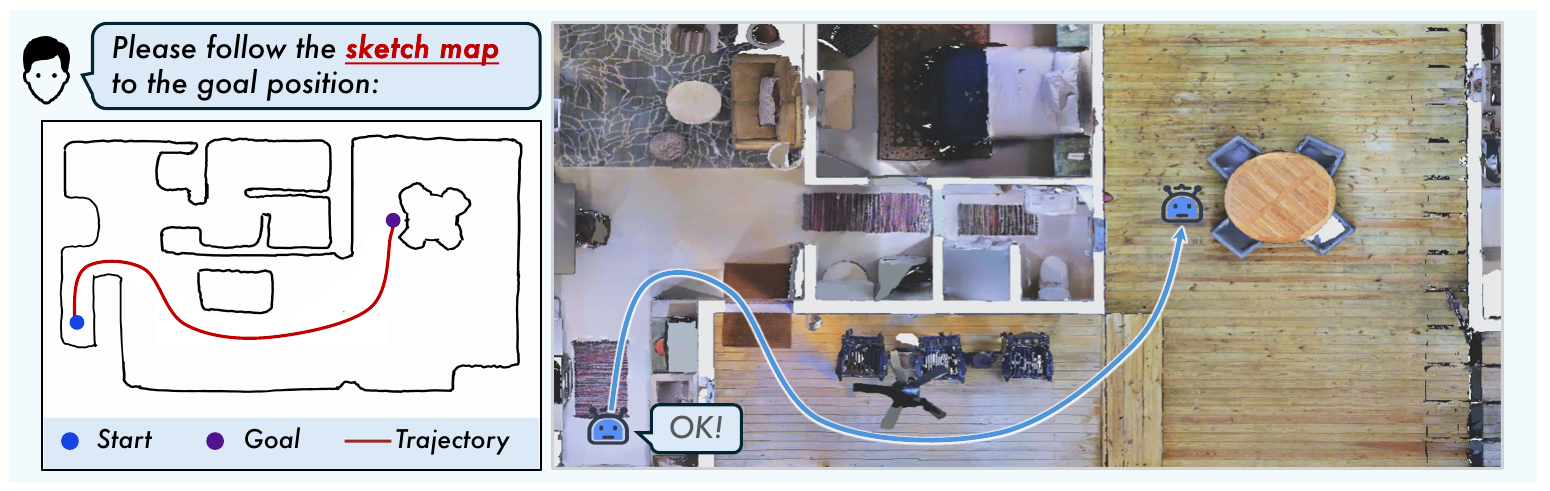}
    \caption{An example of \textbf{Ske}tch map based visual \textbf{Na}vigation (\textbf{SkeNa}). The embodied agent needs to understand the hand-drawn sketch map to reach the goal in a 3D environment with the egocentric observations.}
    \label{fig:fig1}
    \vspace{-1mm}
\end{figure*}

\begin{abstract}
A typical human strategy for giving navigation guidance is to sketch route maps based on the environmental layout. Inspired by this, we introduce \textbf{Ske}tch map-based visual \textbf{Na}vigation (\textbf{SkeNa}), an embodied navigation task in which an agent must reach a goal in an unseen environment using only a hand-drawn sketch map as guidance. 
To support research for SkeNa, we present a large-scale dataset named \textbf{SoR}, comprising 54k trajectory and sketch map pairs across 71 indoor scenes. In SoR, we introduce two navigation validation sets with varying levels of abstraction in hand-drawn sketches, categorized based on their preservation of spatial scales in the environment, to facilitate future research. To construct SoR, we develop an automated sketch-generation pipeline that efficiently converts floor plans into hand-drawn representations. 
To solve SkeNa, we propose \textbf{SkeNavigator}, a navigation framework that aligns visual observations with hand-drawn maps to estimate navigation targets. It employs a \textbf{R}ay-based \textbf{M}ap \textbf{D}escriptor (\textbf{RMD}) to enhance sketch map valid feature representation using equidistant sampling points and boundary distances. To improve alignment with visual observations, a \textbf{D}ual-Map \textbf{A}ligned \textbf{G}oal \textbf{P}redictor (\textbf{DAGP}) leverages the correspondence between sketch map features and on-site constructed exploration map features to predict goal position and guide navigation. SkeNavigator outperforms prior floor plan navigation methods by a large margin, improving SPL on the high-abstract validation set by 105\% relatively.
Our code and dataset will be released. 
\end{abstract}

\section{Introduction}

A fundamental challenge in embodied AI involves developing agents capable of navigating and localizing in unfamiliar environments using diverse human guidance formats~\cite{anderson2018vision}. Consider a typical scenario where a passerby asks for directions to a destination located inside a building. When up-to-date maps are unavailable, purely verbal descriptions and gestures often prove inadequate for navigation guidance. In such cases, hand-drawn sketch maps emerge as an intuitive and practical solution, offering one of the most immediate and accessible forms of spatial communication.
Motivated by this observation, we propose a novel navigation task—\textbf{Ske}tch map-based visual \textbf{Na}vigation (\textbf{SkeNa})—where agents must navigate unseen environments guided solely by hand-drawn sketches. SkeNa provides a robust interface that naturally accommodates spatial uncertainty while enabling rapid user input in dynamic or unmapped environments.

Hand-drawn sketches offer three key advantages as a navigation instruction format:
First, they require neither complex prior information, e.g., GPS localization~\cite{savva2019habitat} and floor plan~\cite{chenF$^3$LocFusionFiltering2025a,li2024flona}, facilitating immediate deployment in unfamiliar environments. 
Second, sketch maps closely align with human spatial memory, enabling users to provide them effortlessly. This eliminates the need to translate environmental layouts into intermediary representations (e.g., verbal commands), thereby significantly reducing human-computer interaction overhead~\cite{zhu2021soon}.
Third, among diverse forms of navigation instructions, sketch maps offer lower ambiguity while aligning with humans' expression habits. By encapsulating the environment's key topological relationships, they ensure precise trajectory information transmission.

To support the study in SkeNa, we introduce a large-scale dataset consisting of 54k trajectories across 71 diverse indoor environments from Matterport3D in~\cite{savva2019habitat}, named \textbf{S}ketch \textbf{o}f \textbf{R}oom (\textbf{SoR}). We also design two distinct validation sets, namely the High-abstraction and Low-abstraction sets, based on the degree to which hand-drawn sketches preserve the real-world layout of environments, which facilitates a comprehensive evaluation of the algorithm's performance across sketches with varying styles.
To build SoR, we propose an automatic pipeline for generating human-like sketch maps that preserves essential geometric relationships while abstracting superfluous details. To enhance the realism of the synthesized sketches, we incorporate a style transfer module that emulates human sketching patterns, thereby maintaining both structural integrity and natural stylistic representation. Our framework provides sketch-based services with two different levels of abstraction for practical deployment.

However, as a distinct form of navigation guidance compared to previous navigation settings, hand-drawn sketches pose the following challenges for the embodied agent to utilize. First, the inherent sparsity of sketches—manifested as extensive blank regions—undermines the efficacy of patch-based feature extraction methods (e.g., CNNs or ViT patch projectors) in capturing semantically meaningful representations. Second, the imprecise nature of sketch-based maps, which often omit non-critical details, simplify structural outlines, and distort obstacle distances, significantly undermines the performance of prior methods~\cite{li2024flona,chenF$^3$LocFusionFiltering2025a} that depend on accurate map inputs.


To address these challenges, we propose \textbf{SkeNavigator}, an end-to-end navigation framework that progressively aligns visual observations with hand-drawn maps to estimate navigation targets. 
To better extract the feature information of hand-drawn maps, we design a \textbf{R}ay-based \textbf{M}ap \textbf{D}escriptor (\textbf{RMD}), which leverages equidistant sampling points and boundary distances within the map. RMD enables each sampled point to capture a broader perceptual range, thereby enhancing the comprehensiveness of the extracted sketch map features. To address the distortion and omission of scene information in sketch maps and improve their alignment with the agent’s visual observations, we propose \textbf{D}ual-Map \textbf{A}ligned \textbf{G}oal \textbf{P}redictor (\textbf{DAGP}). Given the inherent similarity between sketch maps and the agent’s top-down exploration map, DAGP utilizes the correspondence between sketch map features and exploration map features to predict the goal position, thereby guiding the agent’s navigation. Experimental results demonstrate that SkeNavigator achieves significantly superior performance in the SkeNa task compared to previous navigators designed for accurate floor plan maps, with a relative SR improvement of 105\% in the unseen validation set with high-abstract sketch maps.

The main contributions of this work are threefold:
\begin{itemize}
  \item We introduce a new navigation task, SkeNa, which requires an agent to navigate using human-drawn route sketches. To support this task, we construct the SoR dataset, comprising 71 diverse indoor scenes and 54k trajectory-sketch pairs. The dataset includes both automatically generated and manually annotated sketches to ensure diversity and realism.
  
  \item  We present an efficient framework for generating human-like sketch maps from scene layouts and trajectory inputs, with controllable levels of abstraction. This pipeline enables large-scale data augmentation, supporting future research in this domain.

  \item  We propose SkeNavigator, a baseline model for SkeNa that integrates RMD Extraction and DAGP to align exploration maps with sketch maps for goal estimation. In contrast to prior floor-plan-based~\cite{li2024flona} navigation approaches, SkeNavigator achieves significantly higher accuracy on SoR. 
\end{itemize}


\section{Related Works}
\label{sec:related_works}
\begin{figure*}[t]
\centering
\includegraphics[width=.95\textwidth]{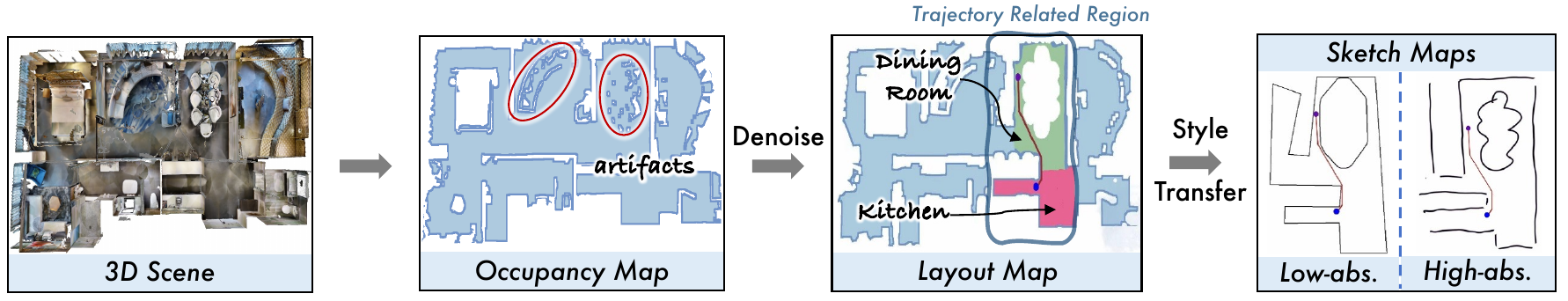}
\caption{\textbf{SoR Collection Pipeline. }Sketch maps with dual-level abstraction are automatically generated by taking 3D scenes as inputs, through constructing layout map, trimming trajectory related region and style transformation.}
\label{fig:sm_clip}
\end{figure*}
\subsection{Visual Navigation}
Visual navigation aims to enable robots or autonomous agents to navigate environments using visual inputs. With advancements in image understanding and natural language processing, a variety of visual navigation tasks have emerged, such as point-goal navigation~\cite{wijmans2019dd}, image-goal navigation, object-goal navigation, and vision-and-language navigation~\cite{ku2020room}.
Point-goal navigation~\cite{ramakrishnan2021habitat,partsey2022mapping} heavily relies on accurate localization and map building, making it less suitable for unknown or dynamically changing environments. Image-goal navigation~\cite{krantz2023navigating,yadav2023ovrl,chaplot2020neural} uses reference images to extract navigational cues, but its generalization capability is limited. Even slight variations in viewpoint or object appearance can cause matching failures. Object-goal navigation~\cite{ramakrishnan2022poni,gadre2023cows,zhou2023esc,yu2023l3mvn} focuses on finding and reaching specific objects but lacks global path planning capability. Vision-and-language navigation~\cite{song2023llm,wang2023dreamwalker} integrates natural language instructions with perception and planning, yet language descriptions often suffer from ambiguity or subjectivity, limiting their effectiveness in visually complex environments.

\subsection{Floor Plan Guided Navigation}
Despite significant advances in these domains, a critical limitation remains the insufficient exploitation of prior structural knowledge about environments. Floor plans, serving as persistent topological representations of spatial configurations, have been extensively employed in navigation studies. 

A line of work focuses on navigation using pre-existing floor plans, which integrates pre-scanned floor plans with sensor data, such as LiDAR~\cite{boniardi2019pose,li2020online,mendez2020sedar,wang2019glfp}, images~\cite{ito2014w,boniardi2019robot,li2024flona}, or visual odometry~\cite{chu2015you}. Some studies further utilize augmented topological maps generated from floor plans~\cite{setalaphruk2003robot} or architectural blueprints~\cite{li2021cognitive} to assist navigation. However, these methods often suffer from environmental mismatches; discrepancies between static floor plans and dynamic real-world layouts can lead to navigation failures, particularly when furniture is moved or layouts are modified. 
Another line of research attempts to construct floor plans during deployment through online mapping and exploration. Some approaches build metrically accurate maps with semantic annotations~\cite{elfes1989using,chaplot2020object,huang2023visual,kwon2023renderable}, while others learn topological graphs for long-horizon planning~\cite{hahn2021no,kim2023topological}. Though these methods adapt well to unknown environments, they typically demand exhaustive exploration, incurring substantial computational overhead and latency. These limitations hinder their deployment in time-critical or resource-constrained applications, such as ad-hoc human-robot collaboration. 
Compared to the previous methods, SkeNa does not provide the precise pre-existing floor plans and does not allow for exhaustive scene exploration. We use a coarsely sketched floor plan as navigation guidance, which is easier for humans to provide and more challenging for embodied agent to follow.
\section{SoR Dataset}
\label{sec:dataset}

To establish a comprehensive benchmarking framework, we propose \textbf{SoR}, a novel large-scale dataset comprising hand-drawn-style maps with corresponding navigation trajectories. SoR contains a total of 54774 sketch-trajectory pairs based on 71 scenes from Matterport3D dataset~\cite{chang2017matterport3d}. 
For evaluation, we design four distinct validation subsets: val-seen/val-unseen scenes with High/Low abstraction levels. The unseen set contains trajectories from scenes not present in the training data, while the high-low abstraction categorization is based on the degree of abstraction in hand-drawn maps. In Section~\ref{sec:data-pipline}, we fully introduce the construction procedure of SoR. Detailed statistics of SoR are provided in the Supplementary Materials.

\subsection{SoR Collection Pipeline}
\label{sec:data-pipline}
As illustrated in Fig.~\ref{fig:sm_clip}, our pipeline comprises three core stages. First, we produce clean path-related environment layouts from the occupancy maps through progressive denoising and cropping. Second,  we extract the trajectory-related region from the environment layout map, ensuring the sketch retains complete coverage of essential areas. Third, the sketch-style transformation module synthesizes two distinct stylized sketch maps from the generated layouts. Subsequently, the automatically generated sketch maps are further subjected to manual verification to ensure quality. 

\subsubsection{Environment Layout Construction} First, an occupancy map for each scene is extracted using the API provided in~\cite{ramakrishnan2021habitat}. Separate occupancy maps are constructed for every floor in multi-story buildings. However, reconstruction artifacts and scattered obstacles often introduce irregular occupied regions into these maps. Since hand-drawn sketches typically abstract away such fine-grained details, retaining these irregularities would yield sketch maps that are either unrealistic or unnecessarily complex. To address this, we apply a denoising pipeline to the raw occupancy maps, producing clean environment layouts for sketch generation. In this pipeline, we first apply morphological erosion to remove isolated obstacles while merging clusters of nearby obstacles into coherent contours. Then we employ binary thresholding to eliminate regions disconnected from the main accessible areas, retaining only the largest connected component as the environment layout.

\subsubsection{Trajectory-Related Region Selection.} We begin by sampling a pair of start and end points on the environment layout map and computing the trajectory between them using the A* algorithm. The trajectory traverses multiple semantically distinct regions (e.g., dining room, kitchen), all of which are preserved in their entirety for sketch map generation to ensure structural completeness. To reduce complexity and adhere to human expression conventions, the remaining parts of the layout map are discarded, as illustrated in Figure~\ref{fig:sm_clip}.
    
\subsubsection{Sketch Style Transformation}To facilitate cross-style generalization, each map is rendered in two distinct abstraction styles:
(1) \textbf{Low-abstraction:} This variant maintains geometric fidelity through polygonal contour approximation, producing structured line drawings that accurately preserve the underlying layout.
(2) \textbf{High-abstraction:} This version simulates free-hand human sketches using a sketch generation method~\cite{vinker2022clipasso} that formulates sketches as collections of B\'ezier curves, while the sketch map's complexity is controlled by adjusting the number of curves. The number of curves is determined by the environment layout's complexity to preserve both the simplicity of sketch maps and the faithful representation of the scene structure. 

\subsubsection{Human Verification}
Finally, we manually review and refine the auto-generated sketches to ensure that trajectories are constrained to plausible walkable regions, as well as line drawings maintain visual clarity without noticeable disruptive artifacts. Besides, whether the layout structure of the occupancy map remains clearly conveyed is also checked. The processed sketches exhibit natural hand-drawing characteristics while preserving interpretability. We also add a subset of sketch maps drawn by humans in the validation set to assess generalization further.

\section{Methodology}
\label{sec:skenavigator}
\subsection{Task Formulation}
\label{sec:task_formulation}
In SkeNa, the agent is initialized with a sketch map $S$ that encodes the environmental layout. This map remains accessible throughout the entire navigation process and provides the agent's initial position and goal location, denoted by blue and purple markers, respectively. At each time step $t$, the agent obtains an egocentric depth observation $D_t$. The agent selects actions from a discrete action space $\mathcal{A} = \{\texttt{STOP}, \texttt{MOVE FORWARD}, \texttt{TURN LEFT}, \texttt{TURN RIGHT}\}$ based on its current state. An episode terminates successfully if the agent executes the $\texttt{STOP}$ action within the maximum allowed steps $T$ and its final position lies within a threshold distance $\tau_d$ of the ground-truth goal location; otherwise, the episode is deemed as a failure.

\subsection{Overall Pipeline of SkeNavigator}

\begin{figure}[t]
    \centering
    \includegraphics[width=0.95\columnwidth]{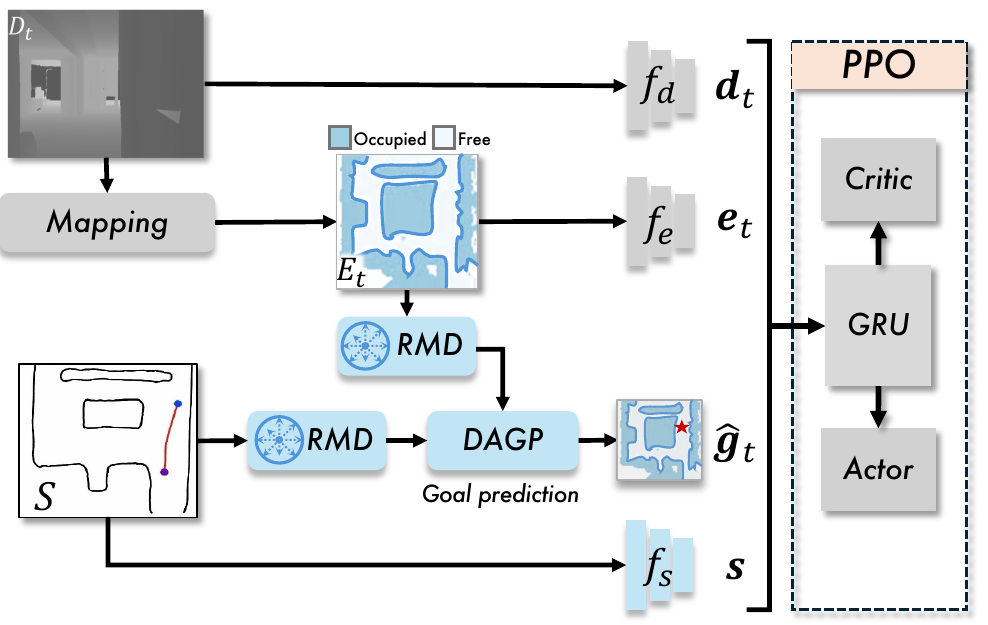}
    \caption{\textbf{Model Architecture.} At time step $t$, Mapping module processes the egocentric depth image \( D_t \) to construct an exploration map \( E_t \). Concurrently, CNN encoders extract features \( \bm{d}_t \), \( \bm{e}_t \), and \( \bm{s} \) from \( D_t \), \( E_t \), and the sketch map \( S \), respectively. These features, along with the goal position \( \hat{\bm{g}}_t \) estimated by DAGP (using RMDs extracted from \( E_t \) and \( S \)), are fed into the policy network to predict actions. 
    }
    \label{fig:main_arc}
\end{figure}

We propose \textbf{SkeNavigator} as the navigation model, whose architecture is illustrated in Fig.~\ref{fig:main_arc}. The model employs a Gated Recurrent Unit (GRU) to maintain the navigation state $\boldsymbol{h}_t$, which is updated at each time step as follows:
\begin{equation}
\boldsymbol{h_t}=\text{GRU}([\boldsymbol{d}_t, \boldsymbol{e}_t, \boldsymbol{s}, \hat{\boldsymbol{g}}_t], \boldsymbol{h}_{t-1}),
\end{equation}
where $\boldsymbol{d}_t$, $\boldsymbol{e}_t$, and $\boldsymbol{s}$ are features extracted from depth inputs $D_t$, exploration map $E_t$, and sketch map $S$, respectively: 
\begin{equation}
\boldsymbol{d}_t=f_d(D_t), \boldsymbol{e}_t = f_e(E_t), \boldsymbol{s} = f_s(S).
\end{equation}
$f_d(\cdot)$, $f_e(\cdot)$, and $f_s(\cdot)$ are CNNs. The exploration map $E_t$ is incrementally constructed and updated from depth observations using the method described in~\cite{zhou2023esc}.
$\hat{\boldsymbol{g}}_t$ is the goal position estimated by the DAGP :
\begin{equation}
\hat{\boldsymbol{g}}_t = \text{DAGP}(S,E_t).
\end{equation}
In DAGP, we employ the RMD to extract features from both \( S \) and \( E_t \), enhancing the representation of layout structures. Details of RMD and DAGP are introduced in Section~\ref{sec:RMD} and Section~\ref{sec:DAGP}, respectively.
SkeNavigator predicts the next action's distribution using:
\begin{equation}
\pi(a_t | \boldsymbol{h}_t) = \text{softmax}(W_{\pi} \boldsymbol{h}_t + b_{\pi}).
\end{equation}
SkeNavigator is trained with Proximal Policy Optimization (PPO)~\cite{ppo}. The training losses are provided in Section~\ref{sec:training}.

\begin{figure*}[t]
	\centering
	\includegraphics[width=0.9\textwidth]{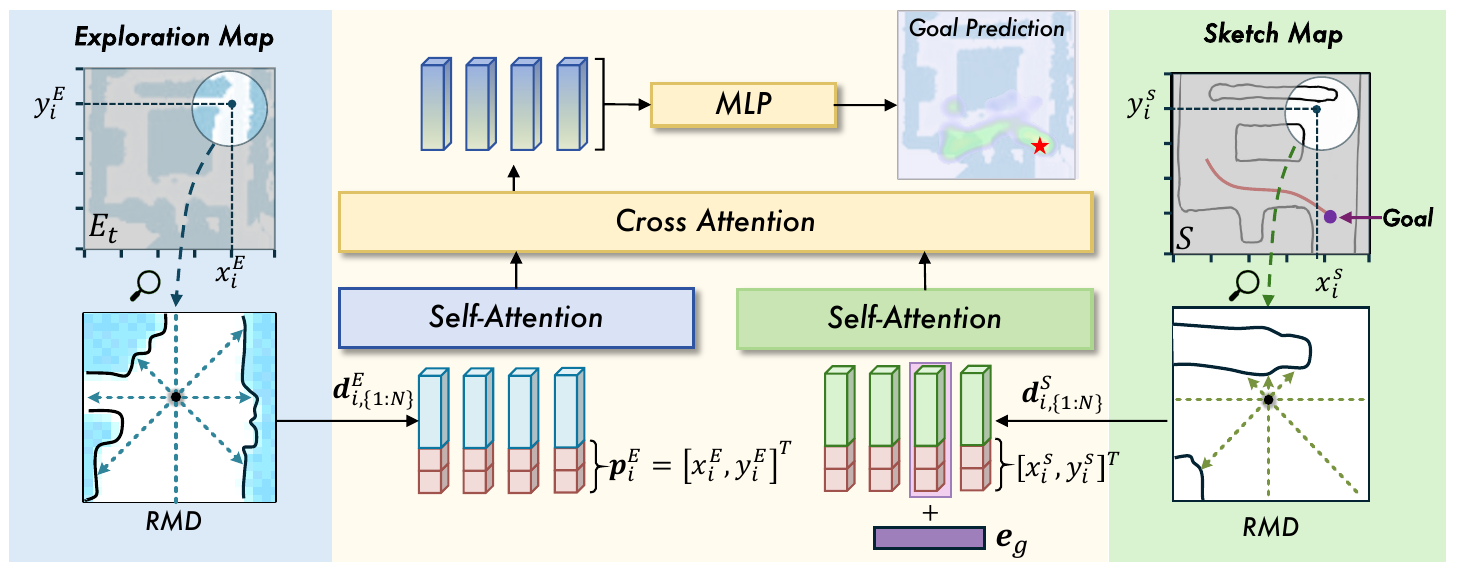}
	\caption{\textbf{Pipeline of RMD Extraction and DAGP.} Illustration of keypoint selection and RMD extraction. Black dots indicate sampled keypoints, with rays encode normalized distances \( d_i^{(j)} \) from each keypoint to the nearest obstacle along the \( j \)-th direction. Together with 2D coordinates $\bm{p}_i^\delta$, after self-attention encoding, a cross-attention block with MLP is imposed and finally output the possibility of each key point being the goal. A learnable embedding is added to the goal-associated keypoint.}
	\label{fig:g_net}
\end{figure*}

\subsection{Ray-based Map Descriptor}
\label{sec:RMD}
Compared to conventional RGB maps, hand-drawn sketches exhibit distinct characteristics, as their primary information is conveyed through sparse strokes. This unique property renders conventional patch-based processing methods (e.g., patch linear projector~\cite{dosovitskiy2020image}) suboptimal, since regions with sparse stroke distributions (\textit{e.g.}, the center of roads) often contain minimal meaningful features in the initial processing stages. To address this limitation, we propose representing each sampled point's features using its obstacle distances from multiple directions, \textit{i.e.}, RMD.
Specifically, we first sample $M=M_1\times M_2$ uniformly distributed keypoints (keypoints are arranged in a structured $M_1\times M_2$ lattice) from each map and casts $N$ uniformly spaced rays around each point, yielding :

\begin{equation}
    \begin{aligned}
            \boldsymbol{R} &= \left[ \boldsymbol{\rho}_{1}, \boldsymbol{\rho}_{2}, \dots, \boldsymbol{\rho}_{M} \right]^{\mathrm{T}} \in \mathbb{R}^{\left( N+2 \right) \times M}, \\
            \boldsymbol{\rho}_{i} &= \left[ d_{i,1}, d_{i,2}, \dots, d_{i,N}, x_{i}, y_{i} \right]^{\mathrm{T}} \in \mathbb{R}^{N+2},
    \end{aligned}
\end{equation}
where $i = 1, 2, \dots, M$ indexes the keypoints, $d_{i,j} \in [0,1]$ denotes the normalized distance from the $i$-th keypoint to the nearest obstacle along the $j$-th ray direction, $\boldsymbol{p}_i = [x_i, y_i]^\mathrm{T}$ represents the keypoint's 2D coordinates. $\bm{R}$ encodes contour information of the local environment at each sampled keypoints, extending to capture obstacle distributions at considerable distances from sampling locations. This representation provides discriminative features that enable effective comparison between exploration maps and hand-drawn sketches within the DAGP framework.


\subsection{Dual-Map Aligned Goal Predictor}
\label{sec:DAGP}

In DAGP, the goal position \( \hat{\bm{g}}_t \) is estimated by leveraging the correspondence between \( S \) and currently \( E_t \).
Consistent with the prior formulation, we extract RMD features: $\bm{R}_t^E = \{\bm{\rho}_i^E\} \in \mathbb{R}^{(N+2) \times M}$ and $\bm{R}^S = \{\bm{\rho}_i^S\} \in \mathbb{R}^{(N+2) \times M}$ from \( E_t \) and \( S \), respectively, using shared hyperparameters. 

To incorporate the destination's positional information on $S$ into \( \bm{R}^S \), we introduce a learnable goal-associated embedding $\boldsymbol{e}_g\in \mathbb{R}^{N+2}$ for the keypoint nearest to the annotated goal position. To find the keypoint, consider
\begin{equation}
    i^*=\mathrm{arg}\underset{i\in \left\{ 1,2,...,M \right\}}{\min}\left\| \boldsymbol{p}_{i}^{\delta}-\boldsymbol{g} \right\| _2
\end{equation}
where $\boldsymbol{g}$ is the goal position marked in $S$. Then we apply
\begin{equation}
    \bm{\rho}_{i^*}^{S}\gets \bm{\rho}_{i^*}^{S}+\boldsymbol{e}_g,
\end{equation}


Before performing feature matching between the two maps, a self-attention block is applied independently to $\bm{R}^S$ and $\bm{R}_t^E$ to contextualize each descriptor within the map’s global structure: 
\begin{equation}
\begin{aligned}
        \hat{\bm{R}}^S &= \mathrm{self\_attn}(\bm{R}^S)\in \mathbb{R}^{M\times d_m},\\
    \hat{\bm{R}}_t^E &= \mathrm{self\_attn}(\bm{R}_t^E)\in \mathbb{R}^{M\times d_m},
\end{aligned}
\end{equation}
where $d_m$ represents the Transformer's output dimension.
To establish geometric correspondence, we apply a cross attention Transformer block by using $\hat{\bm{R}}^E_t$ as queries: 
\begin{equation}
\tilde{\bm{R}}^E_t = \mathrm{cross\_attn}(\hat{\bm{R}}^E_t,\hat{\bm{R}}^S)\in \mathbb{R}^{M\times d_m},
\end{equation}
where $\tilde{\bm{R}}^E_t $ is the fused feature which enables the network to infer semantic correspondences between abstract sketch layouts and explored 3D structure.

$\tilde{\bm{R}}^E_t$ is processed by an MLP to compute $\bm{\omega}$ for the goal's position on the $E_t$: 
\begin{equation}
    \boldsymbol{\omega }=\mathrm{softmax} \left( \mathrm{MLP}\left( \tilde{\bm{R}}^E_t \right) \right) \in \mathbb{R} ^{M \times 1},
\end{equation}
where $\bm{\omega} = \{\omega_i\}_{i=1}^M$ represents the goal's possibility over $M$ keypoints on $E_t$. Each $\omega_i$ corresponds to a 2D coordinates $\boldsymbol{p}_{i}^{E}$ of a keypoint from $E_t$ respectively. 
The predicted goal position $\hat{\boldsymbol{g}}_t$ is obtained through the weighted aggregation of coordinates $\boldsymbol{p}^E_i$: 
\begin{equation}
    \hat{\boldsymbol{g}}_t=\sum_{i=1}^M{\omega _i \cdot \boldsymbol{p}_{i}^{E}}.
\end{equation}
Our method incorporates a soft aggregation mechanism for multiple candidate regions, enhancing robust goal prediction in ambiguous scenarios. Once the goal position is determined, SkeNa reduces to a simplified navigation task with significantly lower complexity.  
\subsection{Training Loss}
\label{sec:training}

SkeNavigator is trained end-to-end through joint optimization of the navigation policy and goal prediction module. The overall loss function is:
\begin{equation}
\mathcal{L}_{\text{total}} = \mathcal{L}_{\text{PPO}} + \lambda_{\text{goal}} \cdot \mathcal{L}_{\text{goal}},
\end{equation}
where $\mathcal{L}_{\text{PPO}}$ denotes the proximal policy optimization loss for navigation, $\mathcal{L}_{\text{goal}}$ represents the $\ell_2$-regression loss for goal prediction, and $\lambda_{\text{goal}}$ controls their relative weighting.

The navigation policy is trained using PPO with a composite reward function defined as:
\begin{equation}
    r_{\text{total}} = \mathbb{I}_d \cdot r_d + \mathbb{I}_i \cdot r_i + \mathbb{I}_s \cdot r_s,
\end{equation}
where $\mathbb{I}_d \in \{0,1\}$ indicate whether the agent reduces the geodesic distance to the goal, $\mathbb{I}_i \in \{0,1\}$ denote if the agent’s action aligns with the expert’s action on the shortest path, and $\mathbb{I}_s \in \{0,1\}$ represent navigation success—the corresponding reward values $r_d$, $r_i$, and $r_s$ weight these terms.
\begin{table*}[!t]
\centering
\begin{tabular}{l l cccc cccc}
\toprule
Dataset & Method & \multicolumn{4}{c}{val-seen} & \multicolumn{4}{c}{val-unseen} \\
\cmidrule(lr){3-6} \cmidrule(lr){7-10}
        &        & SR (\%) & SPL (\%) & SoftSPL & DTG (m) & SR (\%) & SPL (\%) & SoftSPL & DTG (m) \\ 
\midrule
\multirow{4}{*}{L} 
        & Human         & 100 & 88.4 & 88.4 & 0.3  & 100 & 96.1 & 96.1 & 0.16 \\
        \cmidrule(lr){2-10}
        & Random agent  & 0.00 & 0.00 & 0.02 & 9.0 & 0.00 & 0.00 & 0.01 & 8.8 \\
        & FloDiff       & 3.6 & 3.3 & 7.3 & 8.5  & 3.9 & 3.3 & 7.5 & 9.2 \\
        & \textbf{SkeNavigator} & \textbf{17.4} & \textbf{16.6} & \textbf{21.6} & \textbf{6.7} & \textbf{12.7} & \textbf{11.9} & \textbf{15.1} & \textbf{7.8} \\ 
\midrule
\multirow{4}{*}{H}
        & Human          & 96.2 & 81.9 & 85.9 & 1.0 & 94.3 & 82.2 & 86.2 & 0.9 \\
        \cmidrule(lr){2-10}
        & Random agent  & 0.00 & 0.00 & 0.01 & 10.1 & 0.00 & 0.00 & 0.01 & 9.0 \\
        & FloDiff       & 3.5 & 2.6 & 6.2 & 9.9  & 3.9 & 3.8 & 7.8 & 8.7 \\
        & \textbf{SkeNavigator} & \textbf{9.0} & \textbf{7.8} & \textbf{14.3} & \textbf{9.2} & \textbf{8.0} & \textbf{7.6} & \textbf{11.3} & \textbf{8.3} \\
\bottomrule
\end{tabular}
\caption{\textbf{Results.} Performance of different methods. "Dataset" column uses L = Low-abstract, H = High-abstract. }
\label{table:result}
\end{table*}
The DAGP is supervised through L2 loss:
\begin{equation}
    \mathcal{L}_{\text{goal}} = \| \hat{\boldsymbol{g}}_t - \boldsymbol{g}^* \|_2^2
\end{equation}
where $\hat{\boldsymbol{g}}_t$ denotes the predicted goal coordinates and $\boldsymbol{g}^*$ represents the ground-truth location. This objective ensures geometric alignment with the sketch map representation.

\section{Experiments}
\label{sec:experiments}

\subsection{Implementation Details}
We use a GRU with a hidden state size of 512 to model temporal context. The policy is trained with Proximal Policy Optimization (PPO), using a rollout length of 150 steps and four epochs per update. Generalized Advantage Estimation (GAE) is employed with $\gamma = 0.99$ and $\tau = 0.95$. The learning rate and PPO clipping range are linearly decayed throughout training. we set the reward as follows: $r_d$ = 2.0,  $r_i$ = 0.05, and  $r_s$ = 10.0. In DAGP, we first extract RMDs with $M_1=M_2=9$, $N=8$, then we employ a 2-layer attention network. Training is conducted using the AdamW optimizer with a weight decay of 0.01 and an initial learning rate of $10^{-5}$, decayed over epochs. The $\lambda_{\text{goal}}$ is set to 1. 

\subsection{Experiment Setup}

\subsubsection{Metrics}
We adopt four evaluation metrics for SkeNa, \textit{i.e.}, Success Rate (SR)~\cite{anderson2018vision}, Success weighted by Path Length (SPL)~\cite{anderson2018vision}, SoftSPL~\cite{chenOmnidirectionalInformationGathering2023}, Distance to Goal (DTG)~\cite{li2024flona}. Refer to Supplementary Material for details. 

\subsubsection{Baselines}
To validate the effectiveness of SkeNavigator, we compare its performance against the following baselines, under consistent sketch-based input settings:

\begin{itemize}
    \item \textbf{Random agent}: A naive policy in which the agent randomly selects one of the four available actions at each timestep.
    \item \textbf{Human}: Human participants control the agent using sketch map, RGB input, and the on-site exploration map.
    \item \textbf{FloDiff}~\cite{li2024flona}: A recent state-of-the-art method based on diffusion policy learning. We adapt FloDiff to our sketch-based setting by replacing the original floorplan input with human-drawn sketches, while preserving its model architecture and training pipeline.
\end{itemize}


\subsection{Comparison with Baselines}

\begin{table*}[ht]
\centering
\begin{tabular}{
>{\centering\arraybackslash}p{0.9cm}  
>{\centering\arraybackslash}p{0.6cm}  
>{\centering\arraybackslash}p{0.3cm}  
>{\centering\arraybackslash}p{1.0cm}  
>{\centering\arraybackslash}p{1.2cm}  
>{\centering\arraybackslash}p{1cm}    
>{\centering\arraybackslash}p{1cm}    
>{\centering\arraybackslash}p{1.0cm}  
>{\centering\arraybackslash}p{1.0cm}  
>{\centering\arraybackslash}p{1cm}    
>{\centering\arraybackslash}p{1cm}    
>{\centering\arraybackslash}p{1.0cm}  
>{\centering\arraybackslash}p{1.0cm}  
}
\toprule
\multirow{2}{*}{Dataset} 
& \multirow{2}{*}{RGB} 
& \multirow{2}{*}{D} 
& \multirow{2}{*}{Mapping} 
& \multirow{2}{*}{DAGP} 
& \multicolumn{4}{c}{val-seen}
& \multicolumn{4}{c}{val-unseen} \\
\cmidrule(lr){6-9} \cmidrule(lr){10-13}
& & & & & SR(\%) & SPL(\%) & SoftSPL & DTG(m) & SR(\%) & SPL(\%) & SoftSPL & DTG(m) \\
\midrule
\multirow{5}{*}{L} 
&        & \checkmark &        &        & 10.9 & 10.3 & 15.9 & 7.3 & 5.6 & 5.4 & 10.1 & 8.4 \\
& \checkmark & \checkmark &        &       & 9.3 & 9.0 & 14.3 & 7.5 & 5.1 & 4.8 & 11.5 & 8.3 \\
&        & \checkmark & \checkmark &       & 12.1 & 11.6 & 15.3 & 7.4 & 8.6 & 8.4 & 13.1 & 8.3 \\
&        & \checkmark & \checkmark & \checkmark & \textbf{17.4} & \textbf{16.6} & \textbf{21.6} & \textbf{6.7} & \textbf{12.7} & \textbf{11.9} & \textbf{15.1} & \textbf{7.8} \\
\midrule
\multirow{5}{*}{H}
&        & \checkmark &        &        & 3.7 & 3.3 & 8.4 & 9.7  & 1.6 & 1.6 & 4.9 & 8.7 \\
& \checkmark & \checkmark &        &       & 3.7 & 3.6 & 7.4 & 9.8 & 1.9 & 1.7 & 5.0 & 8.6 \\
&        & \checkmark & \checkmark &       & 5.4 & 5.1 & 9.2 & 9.6 & 3.9 & 3.7 & 7.4 & 8.5 \\
&        & \checkmark & \checkmark & \checkmark & \textbf{9.0} & \textbf{7.8} & \textbf{14.3} & \textbf{9.2} & \textbf{8.0} & \textbf{7.6} & \textbf{11.3} & \textbf{8.3} \\
\bottomrule
\end{tabular}
\caption{\textbf{Results of Ablation Studies. }The \checkmark in the "RGB" and "D" columns denote whether RGB or depth is used as egocentric image input.The \checkmark in "Mapping" and "DAGP" indicate module inclusion. }
\label{table:ablation}
\end{table*}
\begin{figure*}
    \centering
    \includegraphics[width=1\linewidth]{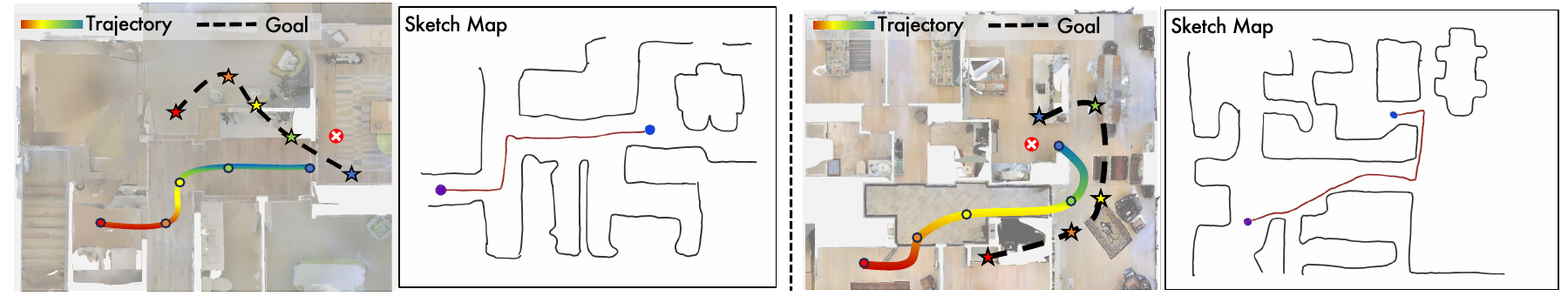}
    \caption{\textbf{Goal Prediction Visualization.}The rainbow-colored curve represents the agent's actual travel path, while the black dashed line indicates the predicted trajectory of the goal point. The circular markers denote the agent's positions at different time steps. Each star indicates the corresponding predicted target point at that time step, and the crosses represent the ground-truth goal positions.}
    \label{fig:vis}
\end{figure*}

To evaluate the effectiveness of our proposed SkeNavigator, we benchmark it against three representative baselines on SoR (Section~\ref{sec:dataset}): a Random policy, agent controlled by human, and a state-of-the-art diffusion-based method for floor plan navigation, FloDiff~\cite{li2024flona}.

 As shown in Table~\ref{table:result} line 1/5, humans achieve SR of over 94\% across all four validation sets in the SoR, demonstrating that the hand-drawn sketches in the SoR dataset are easily comprehensible to humans. In contrast, the random agent yields a success rate of 0\%, indicating that the sampled trajectories in our dataset present considerable challenges.

In Table~\ref{table:result} line 4/8, SkeNavigator achieves consistent and substantial improvements over FloDiff across all key evaluation metrics. Specifically, SkeNavigator attains an average SPL score 11.0\% (Unless otherwise specified, all percentages reported in this section denote absolute differences in numerical values. ) higher than FloDiff on the Low-abstract dataset and 9.0\% higher on the High-abstract dataset, underscoring its superior capability in bridging the gap between abstract sketches and real-world navigation scenarios.

\subsection{Ablation Study}
\label{sec:ablation}

To evaluate the contribution of individual components in our model, we perform ablation studies under five configurations: 
(1) \textit{Depth only}, 
(2) \textit{RGB-D,} 
(3) \textit{Depth + Exploration Map}, 
and (4) \textit{Depth + Exploration Map + DAGP} (our proposed model). The detailed configurations and corresponding experimental results are summarized in Table~\ref{table:ablation}.

\subsubsection{About RGB}

The \textit{RGB-D} modality consistently underperforms the \textit{Depth-Only} baseline across all three validation sets, with the performance gap being particularly pronounced on the Low-abstraction set. Only on the High-abstraction val-unseen subset does \textit{RGB-D} exhibit marginally superior results compared to \textit{Depth-Only}, surpassing the depth-only baseline by a modest 3\% in SR. While the RGB input provides additional visual information, we argue that this modality also introduces texture and color cues. These cues are absent in the human-drawn sketch maps, potentially acting as noise and thereby degrading model performance. 

\subsubsection{About Exploration Map}

Adding the Exploration Map to the depth input yields modest performance gains. The \textit{Depth + Exploration Map} approach consistently outperforms the \textit{Depth only} baseline across all four validation sets, achieving improvements of 1.2\%--3.0\% in SR and 1.3\%--3.0\% in SPL. The map encodes the agent’s visitation history and the layout of free space, which supports local path planning. However, when processed by a standalone CNN encoder, the improvements remain limited. This is likely because CNNs primarily capture local spatial patterns, making it difficult to model long-range dependencies across the map or to align exploration features with the global goal. As a result, the extracted representations lack the structural awareness required for effective goal-directed navigation.





\subsubsection{About DAGP}

Our key component in SkeNavigator, DAGP, significantly outperforms all previous variants. 
DAGP allows the policy to dynamically integrate sketch-derived spatial priors with exploration-driven geometric structure, resulting in substantial gains in SR, SPL and SoftSPL. \textit{Depth + Exploration Map + DAGP} outperforms  \textit{Depth + Exploration Map} at improvement of 3.6\%-5.3\% in SR, 2.7\%-5.0\% in SPL. The performance gap highlights the effectiveness of our approach in fusing dual-map information through structured spatial alignment rather than treating them as separate input streams.

\subsection{Visualization}
As illustrated in Fig. \ref{fig:vis}, we visualize the agent’s trajectory as a color-coded path, along with the corresponding sequence of predicted goal locations. Five discrete time points are marked along the trajectory, with colors transitioning from red to blue to indicate temporal progression. The accuracy of the agent’s goal prediction improves progressively as the explored area expands. Initially, due to limited environmental observations, the predicted goal locations deviate from the ground truth. However, as the agent advances along its path and accumulates more spatial and contextual information, the predictions gradually converge toward the actual goal. This trend demonstrates the agent’s ability to leverage growing observations to refine its goal estimation.

\section{Conclusion}
\label{sec:conclusion}

In summary, this work proposes the SkeNa task, which innovatively employs hand-drawn sketch maps as navigational cues for embodied visual navigation agents. We present an automated pipeline for batch-generating sketch-style maps at various levels of abstraction and construct a large-scale benchmark dataset SoR to facilitate both current and future research. To address the challenges arising from SkeNa, we introduce SkeNavigator. With the RMD extraction and DAGP module, SkeNavigator effectively overcomes the inherent limitations of sketch maps (e.g., imprecise geometry and sparse information) through enhanced geometric feature representation and alignment between sketch maps and the on-site constructed exploration maps. Extensive experiments demonstrate that SkeNavigator significantly outperforms existing baselines, while ablation studies validate the necessity of each proposed component. We hope that this work encourages further research into incorporating abstract human-centric priors into embodied agents.

\bibliography{aaai2026}
\end{document}